\documentclass[journal,11pt,manuscript,onecolumn]{IEEEtran}
\usepackage{amsmath,graphicx}
\usepackage{algorithm,algorithmic}
\usepackage{amsthm,amsmath,amssymb}
\usepackage{color}
\usepackage{graphicx}
\usepackage{subcaption}
\usepackage{wrapfig}
\usepackage{cite}
\usepackage{soul}

\usepackage[margin=1in]{geometry}
\usepackage{setspace}
\title{\huge Signal Processing Challenges and Examples for \textit{in-situ} Transmission Electron Microscopy}

\author{Josh Kacher, Yao Xie, Sven P. Voigt, Shixiang Zhu, Henry Yuchi, Jordan Key, Surya R. Kalidindi}
\begin{document}
%
\maketitle
\begin{abstract}
Transmission Electron Microscopy (TEM) is a powerful tool for imaging material structure and characterizing material chemistry. Recent advances in data collection technology for TEM have enabled high-volume and high-resolution data collection at a microsecond frame rate. Taking advantage of these advances in data collection rates requires the development and application of data processing tools, including image analysis, feature extraction, and streaming data processing techniques. In this paper, we highlight a few areas in materials science that have benefited from combining signal processing and statistical analysis with data collection capabilities in TEM and present a future outlook on opportunities of integrating signal processing with automated TEM data analysis. 
\end{abstract}
%

\section{Introduction}
\label{sec:intro}

Transmission Electron Microscopy (TEM) has long been a powerful tool for imaging material structure and characterizing material chemistry. However, the process to resolve structural features is laborious and time-intensive, drastically limiting the characterization throughput. Recent advances in electron detector technology and computational capacity have facilitated the development of high-speed data collection for TEM with microsecond frame rate acquisition speeds, corresponding to hundreds of gigabytes of data per second \cite{ercius20204d}. This increase in data collection rates presents new opportunities in rapid material characterization and high temporal resolution analysis of transient phenomena. However, it also presents data processing and analysis challenges as the sheer volume of data precludes manual image analysis.
Concurrent with these developments in electron detection, exciting developments in {\it in-situ} technology have been made. It is now possible to physically simulate a wide range of extreme environments and record the material response in real-time. Some of these latest developments include extreme heating rates while maintaining temperature accuracy within one degree, corrosion and other tests in liquid environments on engineering materials \cite{Hayden2019,Mirsaidov2020}, high-temperature gas cells \cite{Li2015}, high cycle fatigue loading \cite{bufford2016high}, local electrical biasing \cite{Gao2014}, and ultrashort reactions characterized using dynamic TEM experiments \cite{Campbell2014}. With increasingly stable {\it in-situ} testing platforms and widespread availability of aberration-corrected instruments, {\it in sit} HRTEM experiments have seen remarkable progress, with examples including direct observations of thermally- or electrochemically-induced phase transformations \cite{Heo2019,Fan2018,Wang2021} and defect interactions captured during mechanical deformation at the atomic scale \cite{Li2021,Wang2018}. Many of the material responses induced by these environments are highly transient in nature, occurring in time frames shorter than can be accurately captured when relying on human reflexes. Often, the operator relies on luck to capture phenomena of interest, limiting the repeatability and statistical significance of the results.

This paper will discuss the development of data processing tools for automated analysis of the microstructure and crystal structure of materials, with an emphasis on applying automated signal processing to understanding and quantifying transient phenomena during {\it in-situ} testing. The motivation of this work is to show how existing signal processing tools can be applied to large data sets generated by modern electron detectors to produce more quantitative, statistically significant results, with an emphasis on {\it in-situ} testing. As the motivation is largely driven by technological advances in electron microscopy and with materials science as the main benefactor, many of the more advanced signal processing methods currently under development will not be discussed. The majority of the examples used to highlight developments and showcase tool applications come from {\it in-situ} TEM liquid cell experiments to investigate corrosion processes in thin metal films at the nanoscale \cite{key2020investigating}. We will highlight a few areas that have benefited from combining signal processing and statistical analysis with new data collection capabilities in TEM and provide motivating applications and illustrative real-data examples.

This paper is organized as follows. Section \ref{sec:overview} includes a brief review of detector development and data acquisition in electron microscopy. Section \ref{sec:image_analysis} contains image analysis techniques for TEM such as segmentation and feature extraction. Section \ref{sec:change-point} presents real-time processing for TEM video sequences using sequential change-point detection. Section \ref{sec:spatial} studies correlations in diffraction space for extracting local crystallographic information. Finally, Section \ref{sec:conclusion} concludes the paper with future outlook in integrating signal processing for automated, high throughput analysis and for establishing rigorous process-structure-property relationships. The examples in this paper primarily involve the analysis of real-space TEM images and diffraction images. We acknowledge exciting developments in areas including scanning TEM (STEM) analysis, such as processing of sparse images, and the integration of high-resolution TEM imaging with {\it in-situ} experiments. However, due to spatial limitations, we have chosen not to focus on these topics. Furthermore, as TEM data sets collected using modern electron detectors are often on the order of terabytes, we focus our remarks on signal processing tools that are computationally inexpensive. This is necessary both for processing large data sets and, with future applications in mind, for in-line analysis during {\it in-situ} experimentation.



\vspace{-0.1in}
\section{Overview}
\label{sec:overview}
\vspace{-0.1in}
\subsection{Signal generation}

\begin{figure}[!h]
\vspace{-0.1in}
\centering
\begin{subfigure}[h]{0.24\linewidth}
\includegraphics[width=\linewidth]{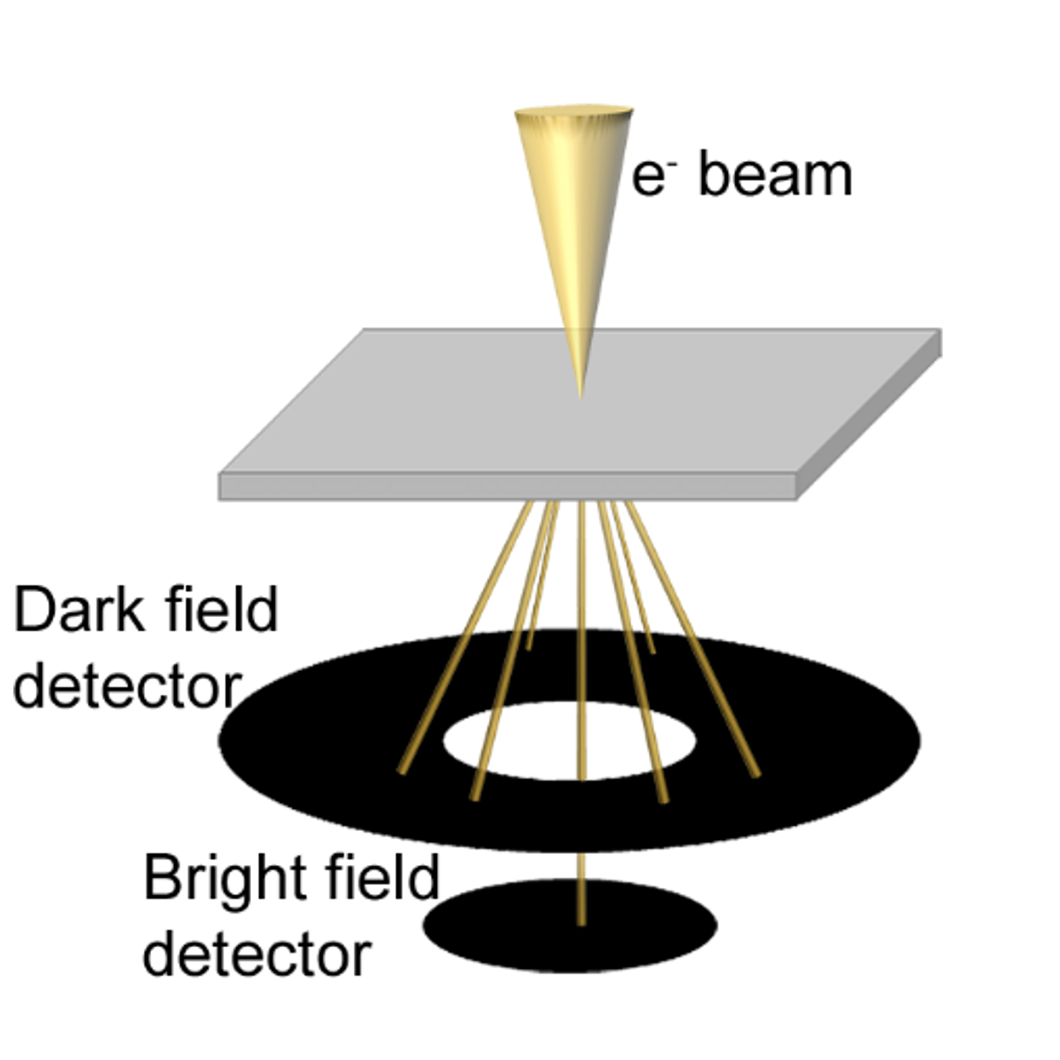}
\caption{}
\end{subfigure}
\begin{subfigure}[h]{0.24\linewidth}
\includegraphics[width=\linewidth]{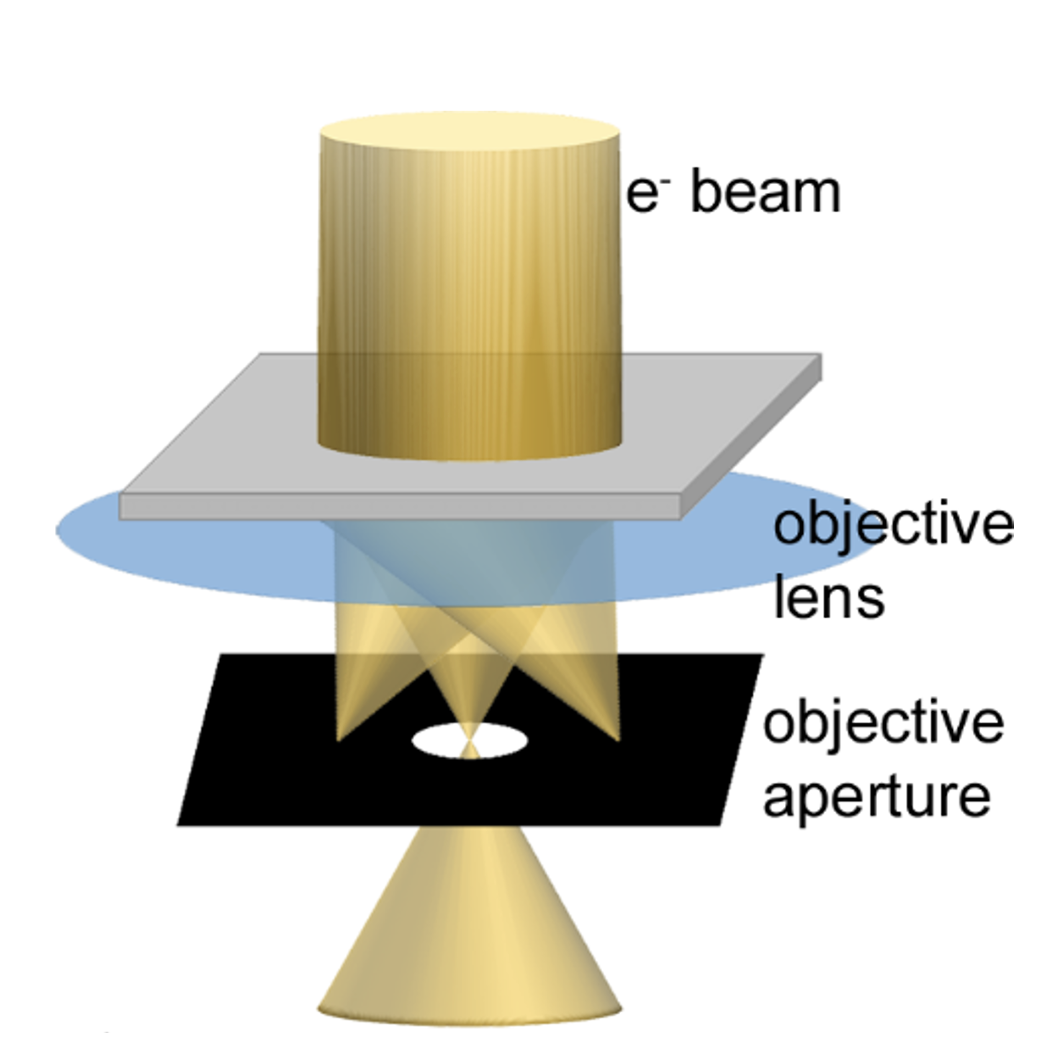}
\caption{}
\end{subfigure}
\begin{subfigure}[h]{0.24\linewidth}
\includegraphics[width=\linewidth]{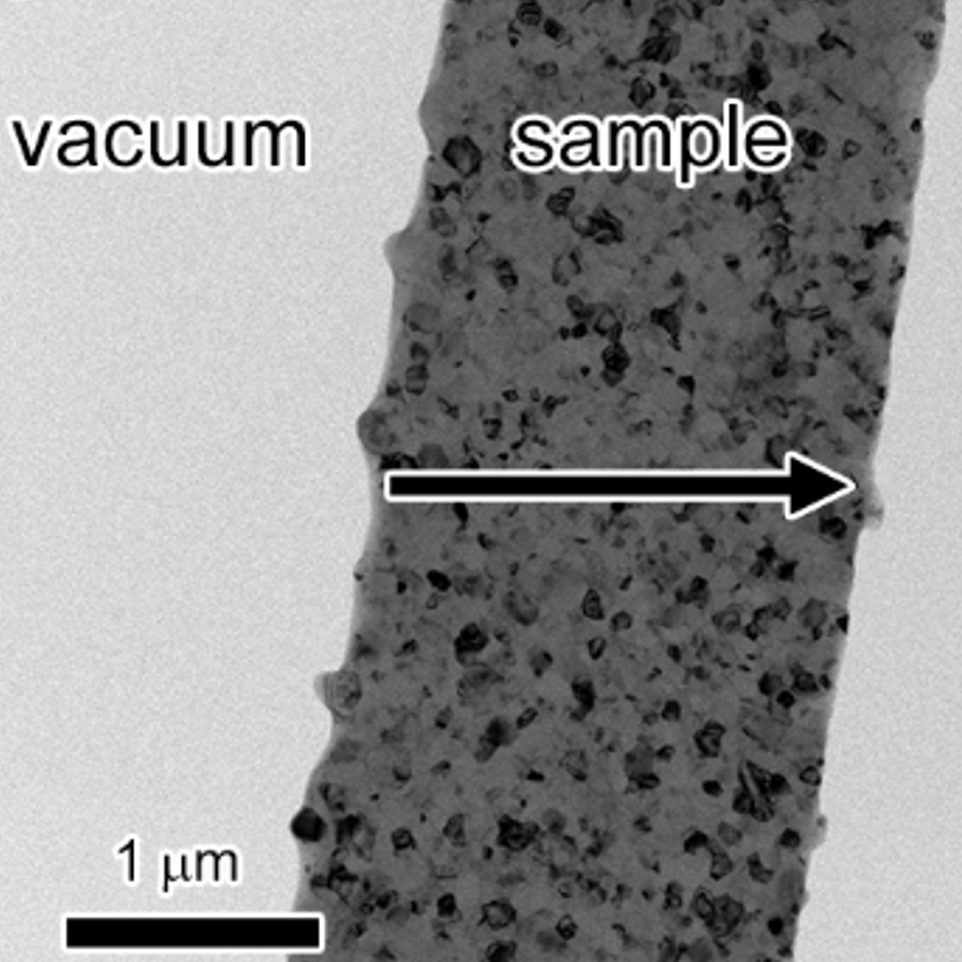}
\caption{}
\end{subfigure}
\begin{subfigure}[h]{0.24\linewidth}
\includegraphics[width=\linewidth]{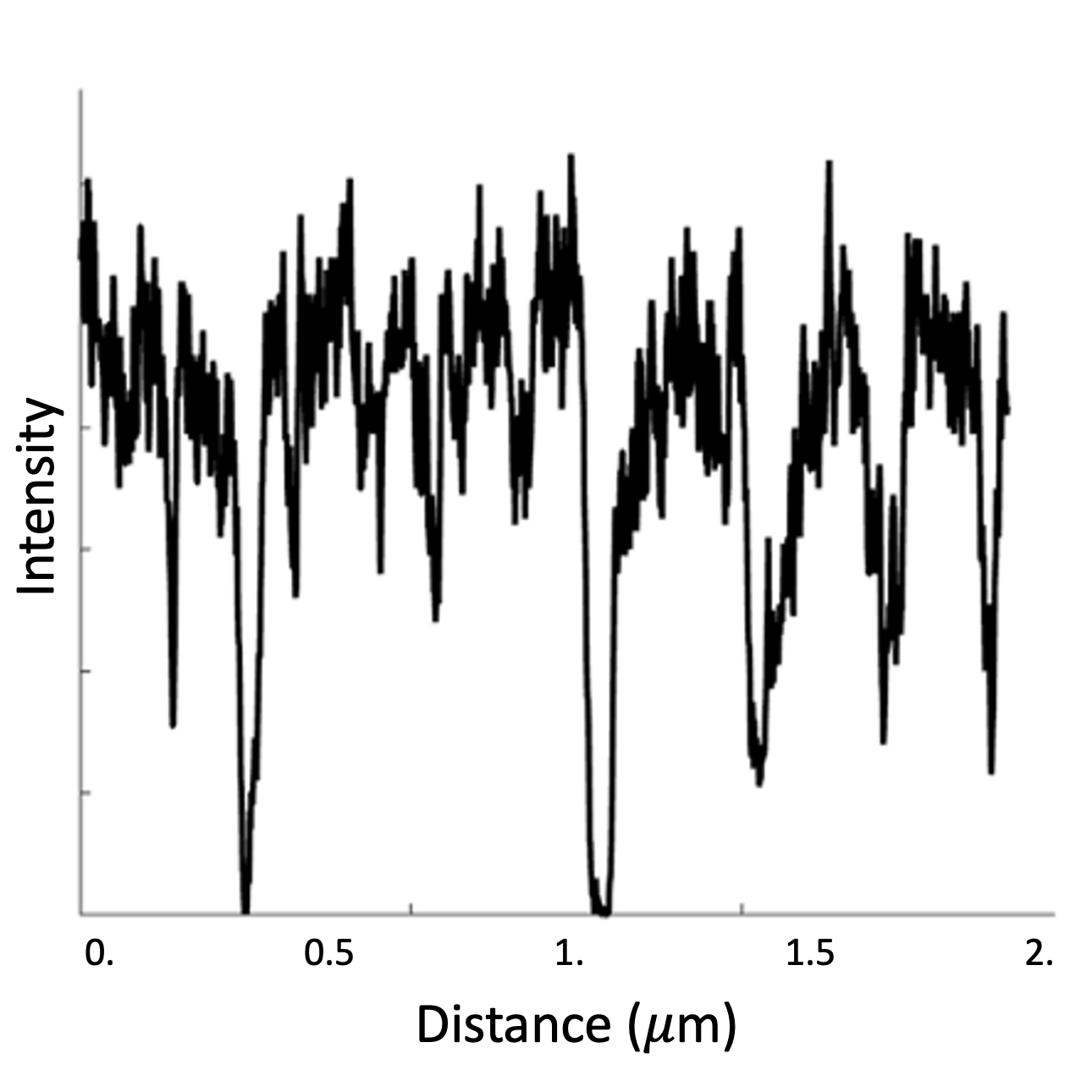}
\caption{}
\end{subfigure}
\caption{Schematic representation of (a) STEM and (b) TEM imaging; (c) example of a bright field TEM image of an Al thin film and (d) intensity profile taken from arrowed line in (c) showing both high and low frequency intensity fluctuations.}
\label{fig:stem}
\end{figure}

In electron microscopy, an image records the total charge carried by electrons to a particular spatial location. In TEM mode, this spatial location corresponds to a location on a detector screen. In scanning TEM (STEM) mode, this spatial location corresponds to the location where the beam converges on the sample. Electrons scattered to different angles contain different types of information, including diffraction contrast (arising from dynamically coupled electron interaction within the sample), mass/thickness contrast, and phase contrast \cite{williams1996transmission}. Depending on which electrons are used to form an image, contrast may arise from a combination of, among other things, crystallographic orientation and strain effects, phase contrast effects, or mass/thickness effects. In both TEM and STEM, the user is able to select which electrons are used to create the final image. In STEM, this is done by varying the camera length and detector used. In contrast, in TEM, this is typically done by selecting a single electron beam using an aperture placed at the focal plane of the objective lens (Figure~\ref{fig:stem} (a)-(b)).

As with all imaging systems, there are a variety of sources of noise in electron microscopy images, including noise in the electronics systems, fluctuations in the electron source, and signal bleeding in the electron detectors. These sources most often contribute white noise to images and data collected during electron imaging. Due to the complex nature of electron/matter interactions, dynamical diffraction contrast effects must be considered while processing electron microscopy signals. Figure~\ref{fig:stem} (c)-(d) shows a bright-field TEM image of a near-uniform thickness pure Al ultrafine-grained thin film. Mass/thickness contrast effects are evident from the higher intensity of the vacuum regions compared to the sample. However, despite its uniformity in terms of chemistry and thickness, both high and low-frequency contrast variations are apparent within the sample itself. The high-frequency variations arise from noise in the detection and illumination system, while the low-frequency variations are from diffraction contrast effects arising from crystallographic orientation changes across the sample. These contrast fluctuations must be considered when extracting data from electron microscopy images.

\vspace{-0.1in}
\subsection{In-situ experiments and data collection} 

Since the initial development of TEM, {\it in-situ} experiments have played an important role in providing real-time insight into the nanoscale response of materials under a variety of different stimuli, including stress, heat, and corrosive environments. This becomes important in resolving ambiguities associated with {\it post mortem} analysis alone. For example, in localized corrosion events,  the scale at which corrosion damage initiates, which is on the order of nanometers, and the rate at which corrosion fronts expand into the surrounding matrix make it difficult to determine corrosion initiation processes using conventional {\it post mortem} characterization. Combining high-resolution capabilities with real-time acquisition can provide direct insight into the mechanism of corrosion damage initiation and the role of microstructural heterogeneities on the susceptibility to local corrosion attack. Understanding this local behavior and how it is influenced by microstructure informs materials sensitive design and contributes to developing superior corrosion-resistant materials.

Due to recent advances in electron detection, {\it in-situ} experiments can now be recorded at microsecond temporal resolution. This is possible due to the transition from charged coupled device (CCD) electron detectors to direct electron detection with complementary metal-oxide-semiconductor (CMOS) detectors. Direct electron detectors do not use a scintillator screen, eliminating the need for fiber optic coupling and reducing the noise in the measurements. This leads to a significant increase in the detection quantum efficiency, enabling low-noise and high frame-rate imaging \cite{Magnan2003,Faruqi2018}. State-of-the-art electron detectors are now capable of collecting thousands of frames per second, with new detectors now capable of imaging rates of tens of thousands of frames per second \cite{ercius20204d}.

\vspace{-0.1in}
\subsection{Process-structure-property relationships}

The ability to collect large amounts of information on the material's internal structure has opened up new research avenues for the development of rigorous mathematical frameworks for the rational design of materials exhibiting a targeted set of materials properties or performance characteristics. Generally referred to as Microstructure-Sensitive Design (MSD) \cite{adams_microstructure-sensitive_2012}, these approaches rely on formulating process-structure-property (PSP) relationships for capturing the core knowledge needed to drive materials innovation efforts efficiently. Modern artificial intelligence/machine learning (AI/ML) tools, customized explicitly for materials research \cite{kalidindi2015hierarchical}, have further enhanced our ability to mine the PSP linkages from available materials data. However, the analysis of microscopy images and the extraction of quantitative and reliable statistical information from the raw microscopy images precedes model building. Therefore, image analysis is a necessary and important tool for advancing the knowledge of materials and their design.     




\vspace{-0.1in}
\section{Image analysis techniques for segmentation and feature extraction}\label{sec:image_analysis}

We will review several critical signal and image analysis techniques for processing TEM data. The examples shown in this section are taken from {\it in-situ} TEM corrosion experiments, with the images collected in bright-field TEM mode, but the tools developed are applicable to a wide range of materials science problems. As outlined in Section II.C., establishing PSP relationships is at the core of materials science. A vital component of this is the ability to rapidly quantify the material response to a stimulus and spatially correlate that response with the local microstructure. This section describes signal and image analysis techniques that allow direct, spatial quantification of material response to a given stimulus using computationally efficient approaches amenable to large data set applications.

Below, we present a few simple segmentation and feature extraction techniques commonly used in the field as illustrative examples. Such relatively simple techniques are appealing and commonly used for TEM data analysis since they are effective in many cases, inexpensive, computationally efficient, and robust in practice, compared with the more sophisticated signal processing techniques. On the other hand, more advanced ML/AI techniques such as deep learning can be potentially developed and applied.

\vspace{-0.2in}
\subsection{Segmentation for real images}
\label{sec:image-segmentation}

Image segmentation is an image analysis procedure that helps extract knowledge from images by assigning semantic labels to all pixels in an image (e.g. ``corroded'' or ``not corroded''). The label is chosen by considering information regarding the pixel, image generation, and the physical specimen. Pixel information includes the pixel's location $x\in\mathbb{R}^d$ in space and time and the pixel's value $p(x) \in \mathbb{R}^F$, where $F$ is a set of dimensions in which the pixel's value may exist. In the simplest cases, the pixel's value is a scalar charge measurement corresponding to the number of electrons detected at a location. However, modern STEM methods allow much more advanced data collection, and a pixel's value can consist of many measurements in many dimensions. For example, electron energy loss spectroscopy (EELS) detection measures energy spectra over the 50-500 meV range for each pixel location \cite{krivanek_vibrational_2014}. STEM imaging has also been used to collect local diffraction patterns, called Ronchigrams, where each pixel's value corresponds to its own two-dimensional diffraction pattern. This imaging technique produces hundreds of Ronchigrams per millisecond at a data collection speed of about 3.2GB/s, and the resulting 4-dimensional measurements take up many hundreds of gigabytes, requiring highly parallelizable analysis techniques \cite{jesse_big_2016}. Information about image generation and the physical specimen cannot be extracted from the image but is based on the real physical conditions that control the observation or measurement of a signal. Some of these include known factors such as chromatic aberrations that can be corrected using hardware solutions in the microscope. Other conditions may not be known. In particular, when considering the application of {\it in-situ} corrosion studies, there is a rapidly changing atmosphere that can interfere with electron beams. Dealing with these unknown conditions is an important part of the segmentation process. Other parts of the segmentation process address the label assignment, the refinement of those labels, and the feature extraction. 

There are many ways to approach the segmentation problem. Some of the current most advanced image segmentation methods depend on deep learning \cite{long_fully_nodate}. However, deep learning is usually not a practical solution to the segmentation of TEM images for two main reasons: (1) they require labeled training data that is usually unavailable for TEM images, and (2) they take much time to train. There is little doubt that deep learning will play a larger role in TEM image analysis in the future, and methods like transfer learning might help make them applicable \cite{decost_uhcsdb_2017}. Some of the challenges will include collecting and sharing segmented datasets \cite{kalidindi_data_2019} and developing deep learning models that can handle TEM images that have memory requirements larger than available GPUs. Currently, unsupervised machine learning methods are used in TEM image analysis. For example, the Ronchigrams have been analyzed by principal component analysis and k-means clustering. However, in this case, there is a need to develop image analysis techniques for large data problems since the Ronchigrams had to be downsampled significantly to allow the analysis on a single CPU, although parallel methods could have enabled the analysis of the complete image. The method was not used for image segmentation since unsupervised classification methods need additional denoising and postprocessing steps to classify pixels and segment images accurately.  A segmentation workflow incorporates a sequence of steps (usually denoising, labeling, then postprocessing).
%
%
In the following three subsections, we will discuss these steps in more detail and present them with a case study as shown in Fig. \ref{fig:Seg_img}.

\subsubsection{Denoising}

Segmentation can be sensitive to noise; in TEM imaging, there are many noise sources such as user input, experimental changes, and instrument noise. The noise is distinct from actual material changes -- the signals-of-interest --but may not be easy to distinguish. For instance, in corrosion applications, there can be a rapidly changing atmosphere environment when the measurements are taken, which will change the brightness of a recorded frame in a video; this introduces a form of high-frequency noise that may vary over time. We may identify the noise's temporal and spatial characteristics in such situations, which can be categorized as either global or local techniques based on the spatial and temporal scale. In particular, global noise affects all pixels in the frame of a video somewhat evenly; moreover, it is more likely to arise due to fluctuations in the gas pressure or thickness of the liquid flowing over samples. On the other hand, local fluctuations affect only a single pixel and arise due to electrons arriving more (or less) in phase with each other at a particular point in space. On the temporal scale level, high-frequency noise usually results from imaging conditions and is unrelated material changes, and thus it is relatively easy to identify and remove. Low-frequency noise is slowly time-varying and may occur on similar time scales as the material's physical changes, making it more difficult to remove.

Extracting the material signal from observations is known as {\it denoising}. Consider a model for the observation at a discrete location indexed by $(n, m)$ in the image: $Y_{n ,m} = X_{n,m} + \epsilon_{n,m}$, where $Y_{n, m}$ is the observation, $X_{n, m}$ corresponds to the material signal and $\epsilon_{n, m}$ is the noise. Generally speaking, there are two approaches for denoising: identifying the signal (based on assumptions and prior knowledge) or identifying the structured noises and then removing them from observation. Assume the noise is normally distributed with zero-mean and variance $\sigma^2$: $\epsilon_{n,m} \sim \mathcal{N} (0, \sigma ^2)$, and thus $\mathbb{E}(Y_{n, m})=X_{n, m}$. Due to a lack of sufficient samples to estimate the expected value, we make an assumption about the {\it local smoothness} property that the pixels $(k,l)$ in  the neighborhood of pixel $(n, m)$, defined with respect to certain distance measure $D(\cdot, \cdot)$ and radius $r >0$:
$\mathcal{K}_{n,m}(r) := \left\{(k,l) | \ D\left( (n,m), (k,l) \right) < r \right\},$ have a similar material signal value, ${X}_{k,l} \approx {X}_{n,m}$. This allows us to estimate the expected value as
\begin{equation} \label{eq:smooth}
    X_{n,m} = \frac{1}{|\mathcal{K}_{n,m}(r)|} \sum_{k,l \in \mathcal{K}_{n,m}(r)} {Y}_{k,l}.
\end{equation}
There are other variants to model noise, such as the bilateral filter \cite{mafi2019comprehensive}. Unlike the average computed in \eqref{eq:smooth}, the bilateral filter computes a weighted average, where weights for pixels in the neighborhood are determined by their distances to center pixel and differences between pixel values. 

When a pixel's value varies over time, it can be corrected using spatial denoising methods since $\epsilon_{n, m}$ is independent of $\epsilon_{k,l}, \forall k \neq n, l \neq m$. However, suppose all the pixel values at a particular time are uniformly changing (as is the case when the atmospheric pressure changes). In that case, temporal information needs to be considered to filter out the noise. The image's brightness can characterize the temporal information over time. For instance, a simple measure of brightness, $B \in \mathbb{R}^T$, is a sequence of averages of frames (of size $M$-by-$N$) in a video of length $T$, $X \in \mathbb{R}^{M \times N \times T}$:
$
    B_t = 1/(MN) \sum_{m=0}^M \sum_{n=0}^N X_{m,n,t}.
$
In some applications, the image brightness may remain constant throughout the measurement of the signal. In such a case, the average image brightness, $\bar{B} \in \mathbb{R}$ can be calculated and the image brightness at time $t$ must be corrected to meet this average, $X_{m,n,t}^* = X_{m,n,t} - B_t + \bar{B}$. This requires knowing a certain functional form of brightness over time. For instance, if the form is linear, then we can use least-squares fit to find the most likely brightness at a time, $\bar{B}_t \in \mathbb{R}^T$. However, in mass/thickness contrast imaging, the amount of corroded area directly determines the image's brightness. Since the corrosion growth rate is the quantity that we aim to estimate from the data, the functional form of $B$ is unknown. In this situation, we may need to infer the brightness $\bar{B}_t$ without making any assumptions on its parametric form. One possibility is to fit a univariate smoothing function, e.g., a spline function, to the data, a piece-wise third-order polynomial that is commonly used in statistical smoothing. The smoothness of the spline is controlled by the maximum second-order derivative at the edges of the intervals. The optimal ``smoothness'' can be determined by performing cross-validation on the regularization parameter \cite{craven1978smoothing,friedman2001elements}. 

Note that many advances have been made in the image analysis field beyond the techniques reviewed above. Such techniques are based on sparse representations (wavelets, dictionary learning), Non-Local (NL) methods (NL-means, NL-Bayes), or to consider the statistical properties of the noises that are non-Gaussian. Such techniques have been are used in many different fields, e.g., for hyperspectral imaging \cite{zhang2015survey} and medical imaging \cite{manjon2008mri}. Adapting such techniques for {\it in-situ} TEM data can be interesting research questions.

%



\subsubsection{Labeling}

Segmentation can produce a labeled image, $\ell \in [1, L]^{M \times N}$, or a labeled video, $\ell \in [1, L]^{M \times N \times T}$. Recall that $M$ and $N$ are the sizes of the image's spatial dimensions, $L$ is the number of labels, and $T$ is the size of the temporal dimension. Labeled videos allow for the extraction of rich quantitative measures and reduce the need for subjective human interpretation \cite{kalidindi2011microstructure,burger2009principles}. Hence, segmentation is ideal for high-throughput imaging since a few key measures can then represent videos.

\begin{figure}[h!]
\centering
\includegraphics[width=.6\linewidth]{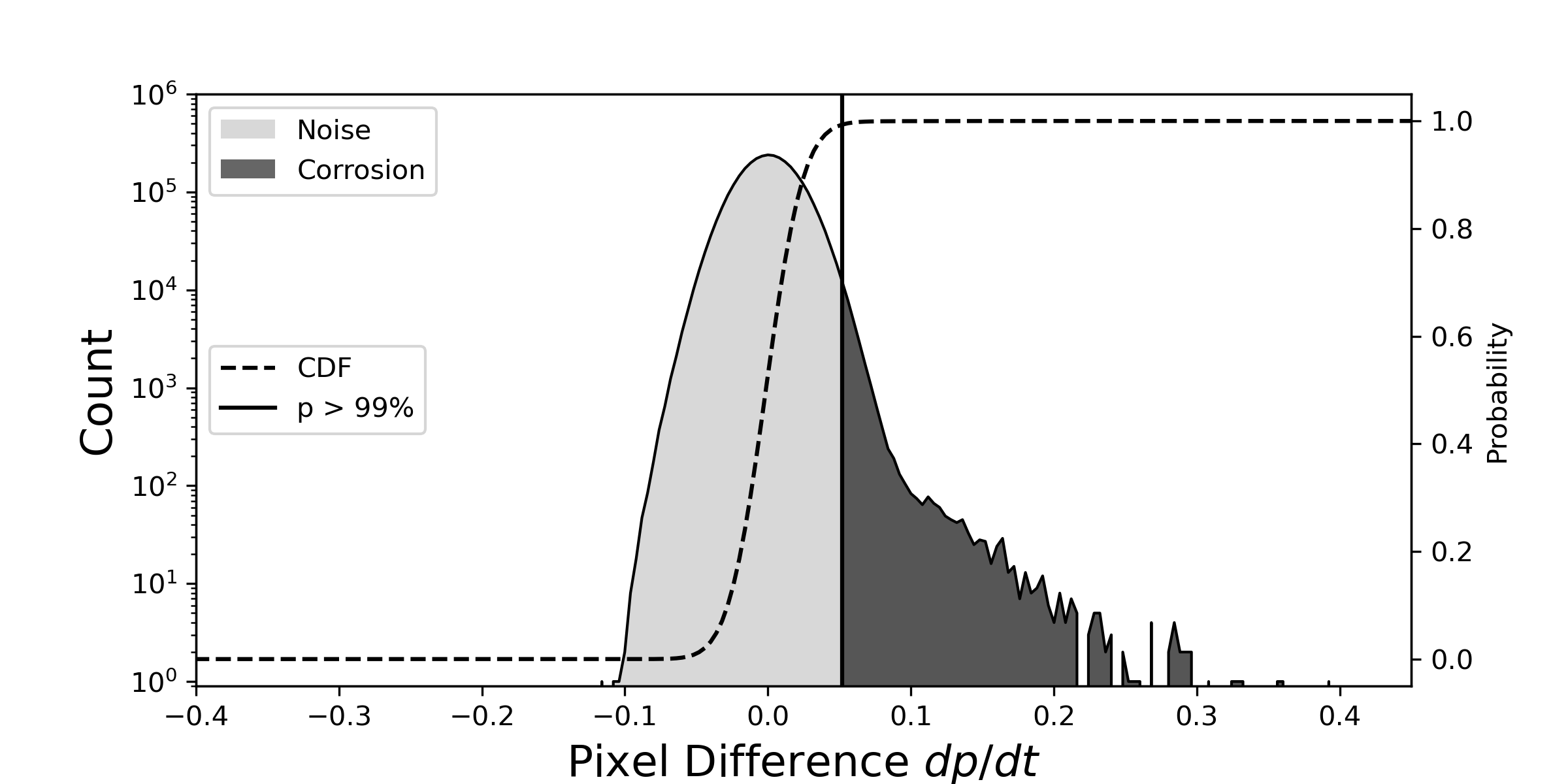}
\caption{Histogram and cumulative distribution function of the forward finite differences of the TEM corrosion video.}
\label{fig:histogram}
\vspace{-0.1in}
\end{figure}

There are a wide array of techniques for labeling pixels in an image, including separating image histograms \cite{otsu1979threshold}, spatial information and graph-cuts \cite{felzenszwalb2004efficient}, and the more advanced end-to-end learned convolutional neural networks \cite{kanezaki2018unsupervised}, and the choice of the method varies case-by-case. For example, in some corrosion studies, we may be only interested in the degradation of the so-called {\it grain boundaries}, which are the boundaries between regions in the microstructure. Labeling the grain boundaries requires edge detection methods. In the case study, the corroded area is brighter than the non-corroded area. Additionally, each pixel begins in an ``uncorroded'' state at $t=0$. This means we are looking for a statistically significant {\it change} in a pixel's value in time considering noise (this can be related to the change-point detection discussed in Section \ref{sec:change-point}). Here we compute the forward finite difference, $\Delta(X)_{m,n,t} := X_{m,n,t+1} - X_{m,n,t}$ and find its histogram. As shown in Fig. \ref{fig:histogram}, the histogram is heavily skewed towards positive pixel changes, indicating that these changes correspond to pixels changing from an ``uncorroded'' state to a ``corroded'' state. Based on this, a pixel switches into a ``corroded'' state at time $t$: $S_{m,n,t}:=1$ if the forward finite difference $\Delta(X)_{m,n,t}$ of the pixel $(m, n)$ at time $t$ exceeds the $99\%$-quantile with respect to the empirical distribution.
%
%
Then we set labels $\ell_{m,n,t}:=1$ when $\sum_{i=0}^t S_{m,n,i} > 0$ to indicate the pixel is ``corroded'', and 0 otherwise as ``uncorroded''.
%
%
Note that $S_{m, n, t}$ highly depends on the difference between noise and actual material changes. Even slight differences, such as those caused by the brightness fluctuations, can significantly impact whether or not we can distinguish between real material changes and image noise. Therefore, correct labeling strongly depends on the noise reduction and filtering techniques in previous steps.

We remark that the segmentation problem here can be related to the signal detection problem (with possibly unbalanced classes), where much research has been done in the signal processing literature. For instance, in signal detection, the theory has been established regarding the fundamental performance trade-off of the detectors between the probability of good detection versus the probability of false alarms. Drawing a precise connection between the two and utilizing existing theory and signal detection techniques for TEM data can be an interesting direction.

\subsubsection{Postprocessing}

\begin{figure}[h!]
\centering

\includegraphics[width=.8\linewidth]{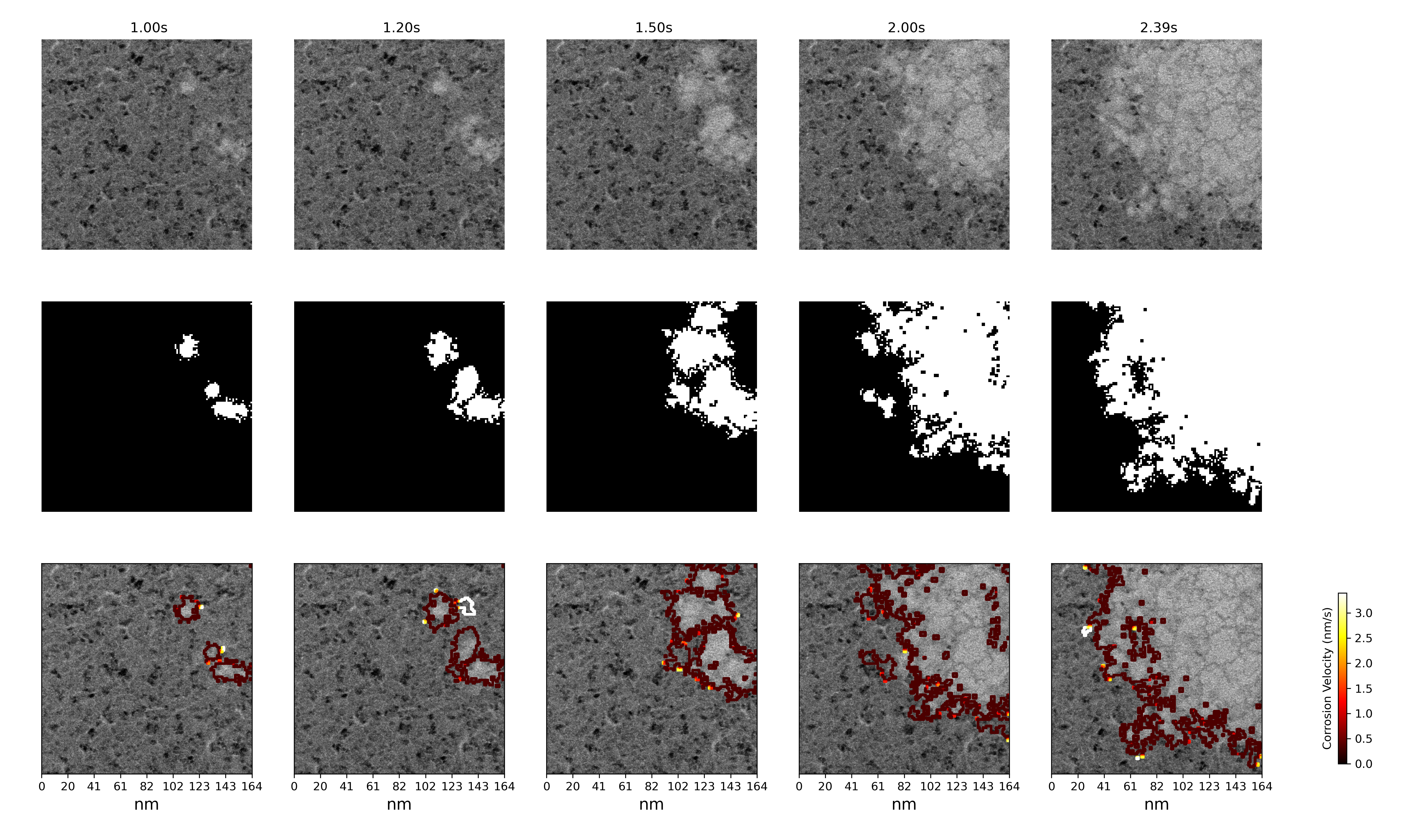}
\caption{Example of image segmentation taken from bright field TEM images collected during an {\it in-situ} TEM corrosion experiment}. Each column is a different frame in the corrosion video. The first row is the original image. The second row is the segmented image. The last row corresponds to fine-grained corrosion velocity.
\label{fig:Seg_img}
\end{figure}

The last step in segmentation is postprocessing, which involves cleaning up the labeled video and extracting features. This step is usually highly problem-specific, and case-by-case solutions can be developed. For instance, a common postprocessing step involves morphological operations such as erosion or dilation, where the boundary of a labeled region is expanded or contracted by a fixed number of pixels. Another commonly used step is morphological smoothing which swaps the pixels' labels if the majority of their neighbors have opposite labels. In the case study, we consider a neighborhood $\mathcal{K}_{n,m}(1)$, and swap the label if it disagrees with more than 50\% of the pixels in the neighborhood; the result is shown in Fig. \ref{fig:Seg_img}.

Feature extraction from the images could be done by assessing global measures such as brightness, which can help us calculate the average corrosion growth rate. However, segmentation enables us to obtain much more fine-grained details, as shown in the last row of Fig. \ref{fig:Seg_img}. For instance, we are interested in the corrosion growth rate at a particular location in physical space, linking microstructural features to the local corrosion growth rate. This may be a critical step in designing microstructures that help reduce the corrosion growth rate and prevent corrosion.



\vspace{-.1in}
\subsection{Sparse feature extraction in diffraction images}
\label{sec:feature-extraction}

In TEM, when the beam of electrons passes through the thin sample, the electrons are treated as waves and diffract in the crystal lattice according to Bragg's Law. That is, the crystal structure of the sample acts as a diffraction grating for the electrons. These diffracted electrons can be captured as diffraction patterns by imaging the focal plane of the objective lens. By acquiring diffraction patterns throughout an {\it in-situ} experiment, changes to the crystal structure, such as phase transformations, precipitation, or amorphization, can be detected. 

\begin{figure}[!t]
\centering
\includegraphics[width=0.90\linewidth]{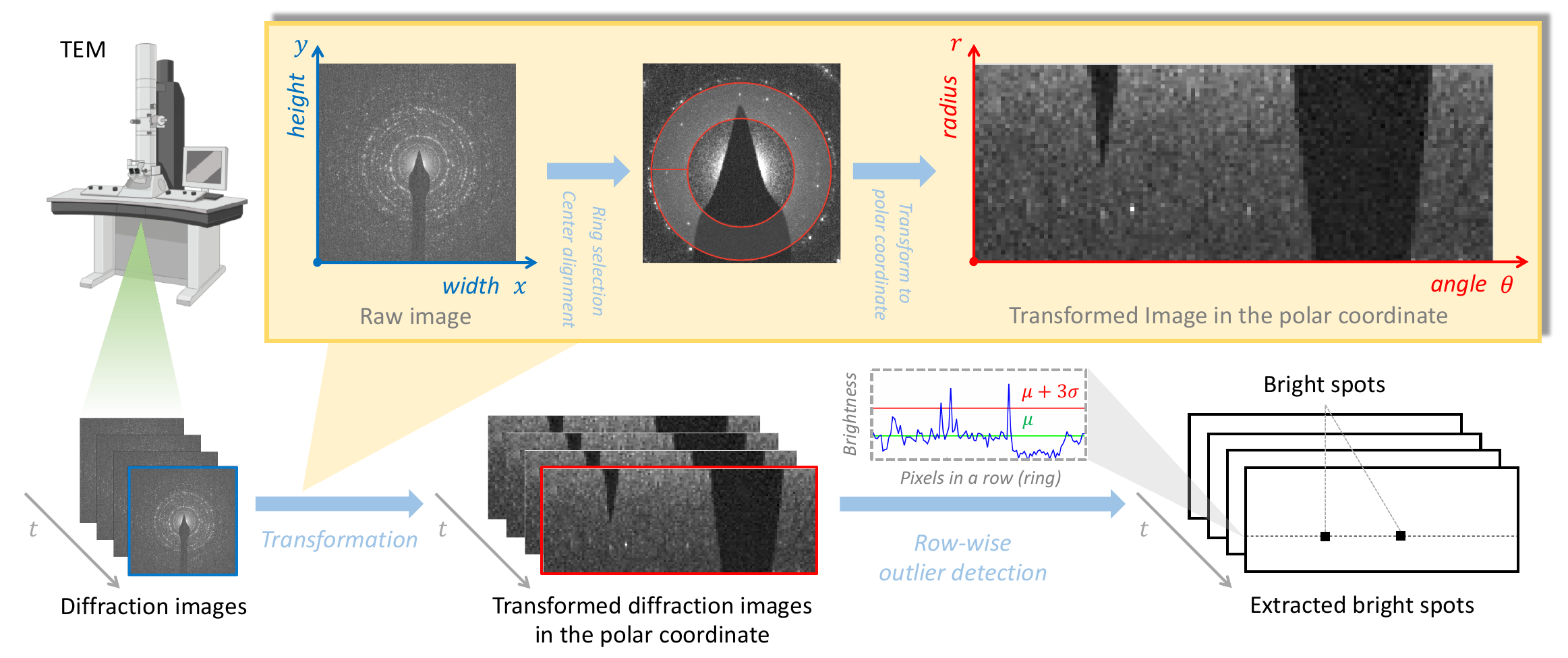}
\vspace{-0.1in}
\caption{An overview of sparse feature extraction in diffraction images. The entire process is composed of two primary steps: (1) transforming the raw diffraction images to the polar coordinate; (2) performing row-wise outlier detection for each image and extracting bright spots. The needle-shaped shadow in the diffraction pattern is a beam stop used to protect the electron detector from the highest intensity electron beams.}
\label{fig:workflow}
\end{figure}

High-intensity local bright spots represent the features of interest in the diffraction image, and they are usually sparse. Fig.~\ref{fig:workflow} shows diffraction patterns collected during {\it in-situ} corrosion of an iron thin film. The collected diffraction patterns contain a series of concentric rings around the center. As most electrons transmit through the sample without interacting, the center of the diffraction pattern is brightest. The high-intensity rings around the center spot arise from diffraction from the iron crystal lattice. This indicates that sparse features exist in rings around the central part of the image. Usually, these spots are pixels with much higher intensities than their local surroundings; however, they may still not be comparable with the pixels in the center of the image. Additionally, relatively low-intensity diffraction spots arising from the formation of iron oxide during the corrosion experiment may exist between the bright diffraction rings. These low-intensity spots provide insight into the corrosion byproducts that form and help us understand the corrosion process in greater detail. This observation, combined with a series of rings that can be identified in the diffraction image, prompts us to transform how the data are visualized. Instead of in the original domain, the nature of the data makes it more appropriate to be studied in the polar coordinate plane.

To detect and identify these sparse features in a diffraction pattern, we carry out \emph{sparse feature extraction}. This procedure aims to remove the background signals in the image because these sparse features are small and primarily exist between the brighter rings. Our focus is, therefore, on the tiny bright spots buried between the brighter rings. A sequence of pre-processing steps has been developed, tailored to emphasize the characteristics of the images. This workflow is illustrated in Fig.~\ref{fig:workflow}. The process consists of two critical steps. The first is the transformation of the diffraction image to polar coordinates, while the second is the bright spot extraction from the image with other known features. 

Below, we illustrate the procedure using a real-data example.
Note that the simple techniques achieved the goal here. Still, more sophisticated feature extraction techniques can be possible -- developing/adapting such techniques for TEM can be an interesting research opportunity that has yet to be realized.

\subsubsection{Diffraction image transformation}

To facilitate extracting the sparse features in the image, namely the bright spots, the polar transformation aims to map the diffraction image to the polar plane. The bright spots can be easily identified in the polar coordinate domain by indexing the pixels with the highest intensity values. The rings seen in the original image are transformed into horizontal stripes, which are easier to process.

There are two primary reasons for conducting this transformation step. First and foremost, a beam stop was used while collecting the diffraction pattern to protect the electron detector, which creates a shadow in the image. This might hamper the feature extraction as it covers some parts of the rings in the image and needs to be carefully removed to prevent erroneously identified features. Additionally, it is challenging to identify the exact location and size of the original image features due to the polar nature of diffraction imaging. For example, suppose there are two features of the same shape and size, but they are presented on two rings with different radius in the diffraction image. Due to this difference in the radius, they will be of different sizes when visualized in the image. The outer ring's feature will take up more pixels than the one on the inner ring, and its shape will be distorted due to the distance from the center. A transformation onto the polar plane can resolve this discrepancy induced in the diffraction.


\subsubsection{Thresholding}

\begin{figure}[!h]
\vspace{-0.1in}
\centering
\includegraphics[width=.5\linewidth]{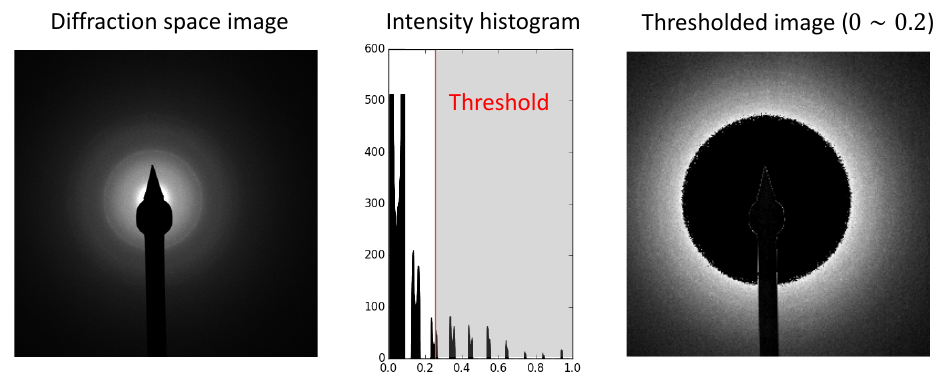}
\includegraphics[width=.45\linewidth]{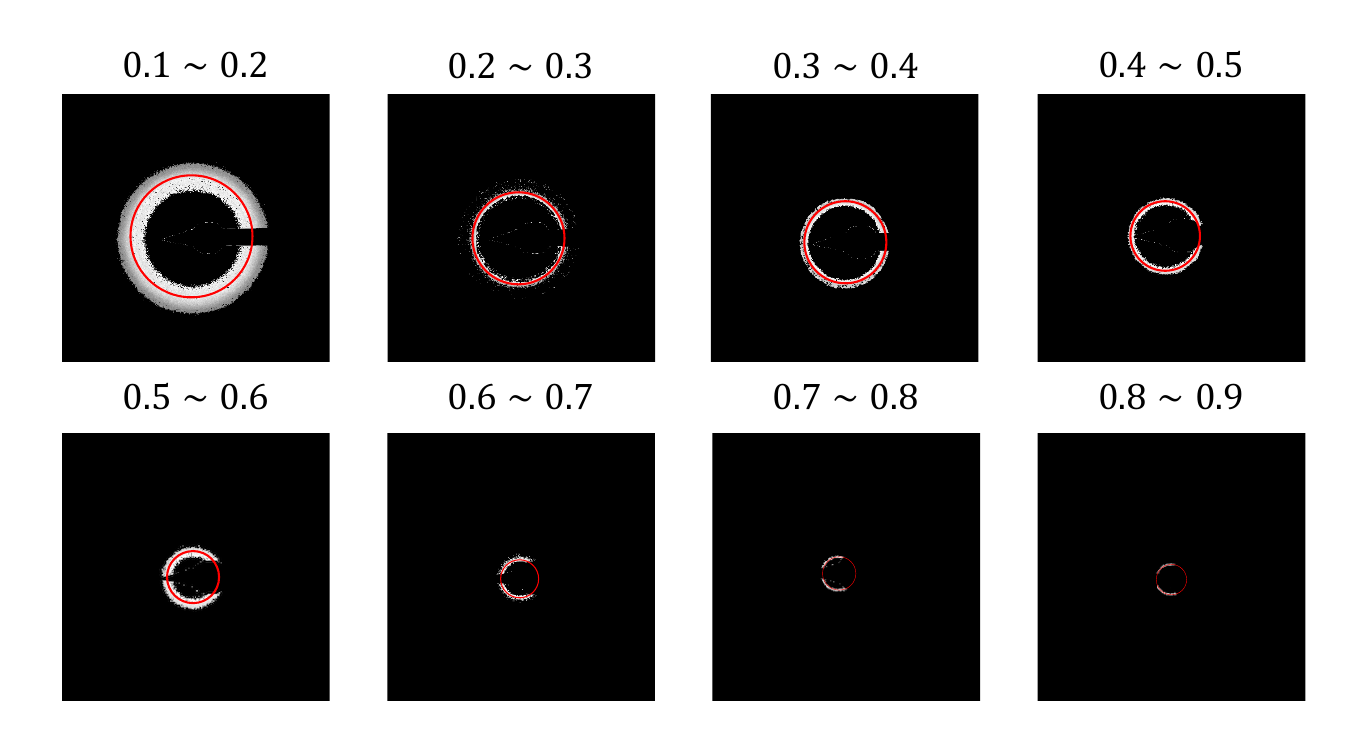}
\caption{Ring extraction in diffusion images. Left: Thresholding to extract image features: a diffraction domain image; the histogram of the intensity; The thresholded image if we
only keep pixels of value in (0, 0.2). Right: Background removal and ring selection: we threshold a diffraction space image with a different range of threshold values, which yields rings at different radii. These concentric rings help to estimate their common center by performing Hough transformation, where the red circles represent the corresponding ring of the estimated center given the non-zero pixels. 
Subsequently, we subtract the irrelevant background and extract the target ring area.}
\label{fig:thresholding}
\end{figure}

To separate the diffraction between the several bright rings, we analyze the histogram of the image pixels shown in the left figure in Fig.~\ref{fig:thresholding} (a), where the histogram of pixel intensities is plotted in the middle. As can be seen, there are several intervals where intensities congregate, with notable gaps in between. For instance, we found that very few pixels in the image have intensities around 0.4, 0.6, and 0.8. This phenomenon conforms to the observation of multiple rings in the image. It indicates that the rings can be separated by placing a threshold on the pixel intensity levels. For example, when the threshold is set around 0.2, the corresponding filtered image is shown in the right of Fig.~\ref{fig:thresholding} (a). We can see the shape of the irregular object is captured by the filtered image. Therefore, if we subtract this from the original image, we will eliminate this unnecessary object. We can further separate the rings from each other by placing thresholds on intensity values subsequently. Overall, thresholding helps remove the undesired object in the diffraction image and successfully separate the several ring components in the visualization for further processing.

\subsubsection{Center alignment} 

Another problem is to align images taken at different times during the experiment. To identify the center of each image taken, we adopt circle Hough transformation \cite{vc1962method} to the thresholded images with different ring structures shown in Fig.~\ref{fig:thresholding} (b). The circle Hough transformation is conducted with the following steps: (1) Define the parameter space of $(x,y,r)$, where $(x,y)$ are coordinates of the center, and $r$ is the radius of the circle. Assign each point in the parameter space with an initial value of zero. (2) Pre-process the image with Gaussian blurring and Canny edge detector. To find the largest visible rings, we apply the Canny edge detection algorithm \cite{canny1986computational} to identify the boundary between the dark and bright areas accurately. (3) For each possible combination of parameters spanned by a circle with $(x,y,r)$, add 1 to their value. (4) The point $(\hat{x},\hat{y},\hat{r})$ with maximum value in the end becomes the circle found. With the center of the rings found in each image, we then align these images with the centers' coordinates.


\subsubsection{Transforming the polar coordinates}

As discussed, feature extraction and the distortion of the images' features motivate the transformation into polar coordinates. The transformation procedure occurs after the center of the rings in the image is found by Hough transformation. To convert the image into the polar plane, we set the center's coordinates as the origin. An angle mask $\theta$ and a radius mask $r$ are applied to the original diffraction image given the circles' center. This masked portion of the image is then the pixel with coordinates $(\theta,r)$ in the polar plane. Through this scheme, we can obtain the transformed image, as shown in the right panel of Fig.~\ref{fig:polar-coord-transform}.

However, it is noted that rings at different radius $r$ contain a different number of pixels in the original image. For instance, the polar coordinate $(r=10,\theta \in [0, 360^\circ])$ contains more pixels than that of $(r=5, \theta \in [0, 360^\circ])$ since the arc is larger. When this happens, we have elected to aggregate these pixels by taking their average intensities in the polar plane. Therefore, there is a trade-off determining the resolution of the angle and the radius masks. The higher resolution might cause one bright spot in the original image to correspond to multiple pixels in the transformed polar image, scattering the feature itself. On the other hand, the lower resolution might result in multiple bright spots close to the center cramped into one single pixel, losing the information. Therefore, the resolution in $(\theta,r)$ should be carefully tuned to ensure the polar image's quality.


\subsubsection{Bright spot extraction}

\begin{figure}[!t]
\centering
\includegraphics[width=.9\linewidth]{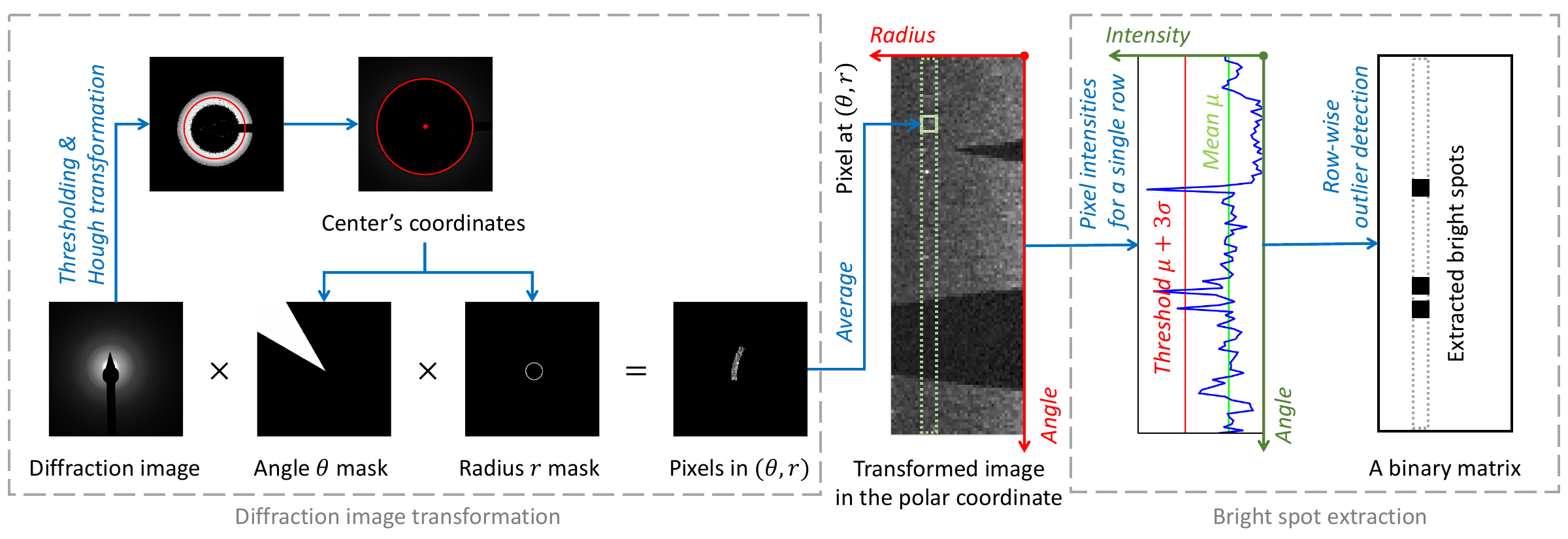}
\vspace{-0.1in}
\caption{An illustration of the process of diffraction image transformation and bright spot extraction. The left panel shows the diffraction image transformation for a single pixel of the transformed image in the polar coordinate; The right panel shows the bright spot extraction for a single row of the transformed image. }
\label{fig:polar-coord-transform}
\end{figure}

The majority of the diffraction image's bright spots do not have steep contrast against its nearby pixels; This makes it challenging to detect by the human eye. Fortunately, it is known from domain knowledge that the bright spots usually appear in the vicinity of a particular radius (though at an unknown angle). Therefore, we can pin down the bright spot around the specified radius. In this step, we focus on the pixels with a specified radius $r$ that is slightly larger than the radius of the ring. Then we formulate the detection objective as a sparse mean shift in the time series intensity data. As shown in the left panel of Fig.~\ref{fig:polar-coord-transform}, we can obtain a $360$-dimensional signal for each transformed polar image that represents the mean pixel intensities across a range of angles and radius values. This step is where we move from visualization to actual signals. We use the processed and transformed image to generate time-series signal data for further change detection analysis.

\subsection{Supervised learning for labeled data}

Thus far, we have focused on unsupervised methods since it is typically expensive to hand label a large amount of TEM data. However, we may also deploy supervised learning methods for TEM data in scenarios where some labels are available. For example, the image segmentation in Section~\ref{sec:image-segmentation} can be considered as a two-class classification problem, in which we aim to identify if a certain area of the material is corroded or not corroded.
For such purpose, we may consider various state-of-the-art supervised machine-learning techniques, such as the support vector machine (SVM), random forests, logistic regression, and deep learning (see \cite{kotsiantis2007supervised} for a detailed review). 
Another example is the signal detection problem in Section~\ref{sec:feature-extraction}, which can potentially be regarded as a hypothesis testing (or signal detection) problem by modeling the data distributions under the two hypotheses \cite{poor2013introduction}. The machine learning techniques tend to have good performance when there is a large amount of labeled data; however, they can be less interpretable (in particular deep learning techniques). On the other hand, the probabilistic model-based statistical signal detection techniques can be more interpretable, and some have the theoretical foundation and performance guarantees.



\vspace{-0.1in}
\section{Real-time processing: Example sequential change-point detection}
\label{sec:change-point}

Advances in TEM technology have enabled new paradigms for sequential and real-time data collection, and thus {\it in-situ} processing of the real-time collected data to detect emerging features becomes a highly desired property for the new TEM system. Currently, the data are captured in real-time but analyzed offline, limiting the experimenter's ability to explore in detail regions of interest while at the microscope. 

Real-time processing streaming data include a wide range of signal processing techniques (see, e.g., a tutorial in \cite{liu2014survey}). There are new challenges and opportunities provided by the new TEM instruments to perform in-situ and/or sequential data acquisitions \cite{taheri2016current}. For instance, when sequential acquisitions are performed, computational imaging problems include registration, drift mitigation, superresolution, deblurring, and particle tracking. To date, there is limited development in signal processing techniques as described above for in-situ TEM.

Here, we focus on a specific real-time processing technique: online change detection as an illustration. Sequential change detection is an important technique for in-situ TEM as it provides automatic and unsupervised labeling of the data. The signal detection task in TEM has two characteristics. First, each observation is a very high-dimensional vector, so we need to develop an algorithm that can sequentially handle a large amount of data. Second, the change is \textit{sparse} in the sense that among all the parameters, only a small proportion of the changes after the unknown change-point. Sequential adaptive change detection method for {\it in-situ} TEM signal detection has been developed in \cite{cao2018sequential, caozhu2018sequential} based on advances in statistical sequential (real-time) change-point detection (see a general survey \cite{xie2021sequential}). The algorithm is developed by assuming Gaussian observations and the signal being a sparse mean shift. The method can precisely control false alarms and be computed recursively, thus automating the detection in real-time.

In this section, we present our two sequential adaptive detection algorithms that can process images in real-time and detect change points that mark the appearance of salient features in the videos.

\vspace{-0.1in}
\subsection{Change-point detection problem set-up}

Assume a sequence of $d$-dimensional observations $X_1, X_2$, $\ldots$ which are i.i.d. random variables from a multivariate normal distribution $\mathcal{N}(\theta, I_d)$ with {\it unknown} mean parameter $\theta \in \mathbb{R}^d$ and $I_d$ denotes a $d$-by-$d$ identity matrix. We will estimate $\theta$ online to be adaptive. 
Consider the sequential change-point detection problem that the underlying distribution of the data changes from a known state to an unknown state after at an unknown change-point $\nu$. Without loss of generality, we assume that the pre-change mean is an all-zero vector. The post-change mean is unknown and belong a set $\mathcal{A}$ defined as $\mathcal{A} = \{\theta: \|\theta\|_0 \leq s \}$, where $\|\cdot\|_0$ is the number of non-zero entries of $\theta$ and $s$ is a prescribed value to characterize the sparsity. Formally, we consider the following sequential hypothesis test: 
\begin{equation}
\begin{split}
\textsf{H}_0: &~~X_1, X_2, \ldots \overset{\rm i.i.d.}{\sim} \mathcal{N}(0,I_d), \\ 
\textsf{H}_1: &~~X_1, \ldots, X_{\nu} \overset{\rm i.i.d.}{\sim}\mathcal{N}(0,I_d), \quad X_{\nu+1}, X_{\nu+2}, \ldots \overset{\rm i.i.d.}{\sim} \mathcal{N}(\theta,I_d), ~~\theta \in \mathcal{A},
\end{split}
\label{maintestproblem}
\end{equation}
where i.i.d. means independent and identically distributed. The goal is to detect the change as quickly as possible after it occurs under the false alarm constraint. 
We will consider likelihood ratio-based detection procedures that we call the adaptive CUSUM (ACM) and adaptive SRRS (ASR) procedures. 
\vspace{-0.1in}
\subsection{Online change-point detection procedures}

Now we derive the detection statistics. For each putative change-point location $k$ before the current time $t$, the post-change samples are $\{X_{k}, \ldots, X_t\}$, and the post-change parameter is estimated as 
$
\hat{\theta}_{k,i} = \hat{\theta}_{k,i}(X_k, \ldots, X_{i}), \quad i\geq k.
$
Denote $f_{\theta}$ as the density function for $\mathcal{N}(\theta, I_d)$. The likelihood ratio at time $t$ for a hypothetical change-point location $k$ is given by (initialized with $\hat{\theta}_{k,k-1} = \theta_0$)
\begin{equation}
\Lambda_{k,t} = \prod_{i=k}^t  \frac{f_{\hat{\theta}_{k,i-1}}(X_i)}{f_{0}(X_i)}, 
\label{cumulativestat}
\end{equation}
where $\Lambda_{k, t}$ can be computed recursively since
\[
\Lambda_{k,t} = \Lambda_{k,t-1} \cdot \frac{f_{\hat{\theta}_{k,t-1}}(X_t)}{f_{0}(X_t)}.
\]

Since the change-point location $\nu$ is unknown, due to the maximum likelihood principle, we take the maximum of the statistics over all possible values of $k$. We consider window-limited versions \cite{willsky1976generalized} to avoid infinite memory, by taking the maximum over $k\in [t-w,t]$, where $w$ is a prescribed window size. This leads to the ACM procedure 
\begin{equation}
T_{\rm ACM}(b) = \inf \left\{t\geq 1: \max_{t-w\leq k\leq t} \log \Lambda_{k,t} > b\right\}, 
\label{ACMprocedure}
\end{equation}
where $b$ is a pre-specified threshold to control the false-alarm rate. 
The Shiryaev-Roberts (SR) procedure is defined similarly by replacing the maximization over $k$ in (\ref{ACMprocedure}) with the summation, which can be justified from a Bayesian prior assumption.
As shown in \cite{cao2018sequential}, the performance of the ACM and the ASR is very similar. However, the likelihood ratio in (\ref{cumulativestat}) can explode when $d$ is very large. Thus, in practice we prefer to use the ACM procedure to avoid possible numerical issues. 

\vspace{-0.1in}
\subsection{Computationally efficient online parameter estimation}

The detection statistic requires a sequence $\{\hat{\theta}_{k,t}\}$ of estimators 
from streaming data. In practice, to achieve the computational efficiency required by the online algorithm, we construct a sequence of estimators using online mirror descent (OMD). The main idea of OMD is that, at each time step, for any $k$, the estimator $\hat{\theta}_{k,t-1}$ is updated using the new sample $X_t$, by balancing the tendency to stay close to the previous estimate against the tendency to move in the direction of the greatest local decrease of the loss function. The advantages of OMD are (1) it allows a simple {\it one-sample update}: the update from $\hat{\theta}_{k,t-1}$ to $\hat{\theta}_{k,t}$ only uses the current sample $X_t$, and the update for the detection statistic has a simple recursive scheme. This is the main difference from the traditional generalized likelihood ratio statistic \cite{lai1998information} where each $\hat{\theta}_{k,t}$ is the exact maximum likelihood estimate obtained using all the historical samples. (2) OMD is a generic algorithm for solving the online convex optimization (OCO) problem \cite{hazan2016introduction}. 
In \cite{cao2018sequential}, it is proven that even these approximate maximum likelihood estimation schemes have minimal statistical efficiency. 

\vspace{-0.1in}
\subsection{Real-data example}

\begin{wrapfigure}{r}{0.4\textwidth}
\centering
\vspace{-0.2in}
\includegraphics[width=1.\linewidth]{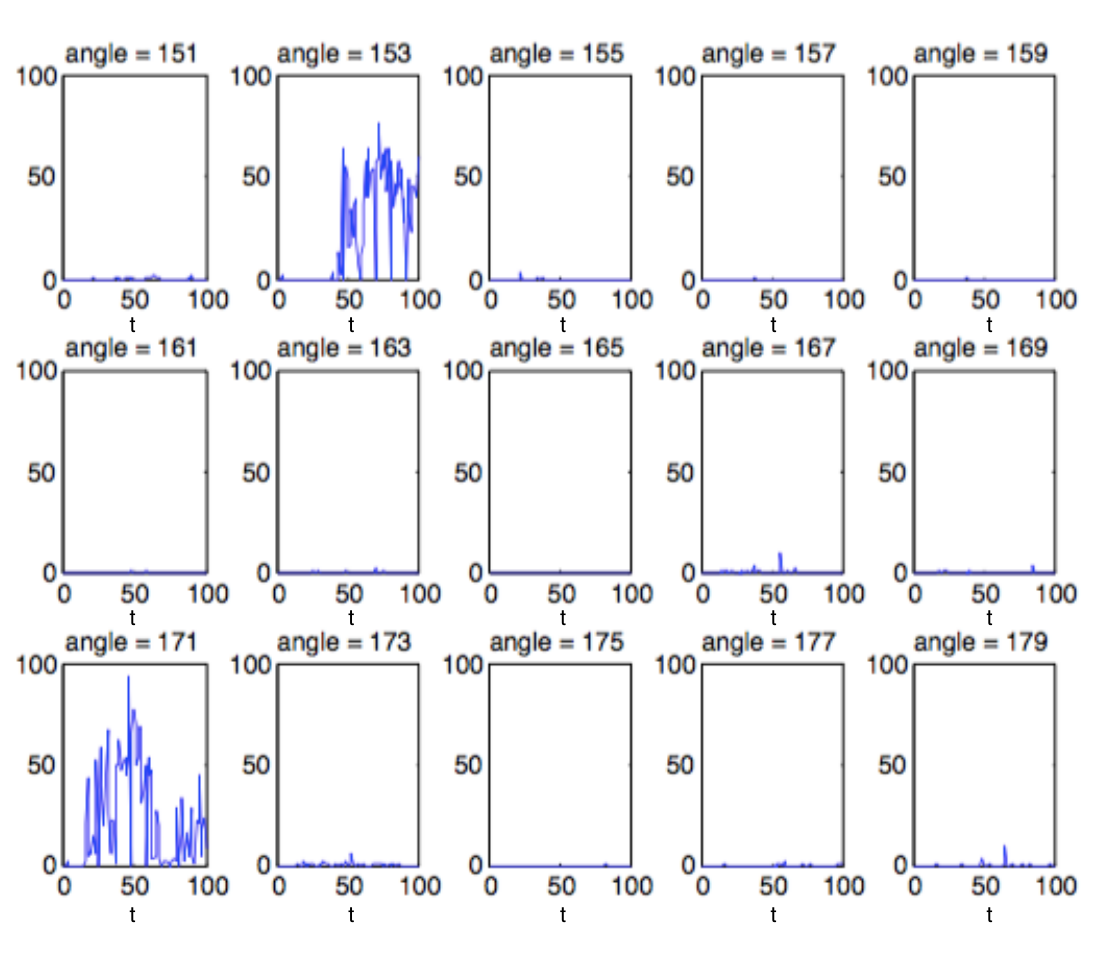}
\caption{The extracted signals for the sequence of $100$ images (for a subset of selected angles). For each sub-figure, the horizontal coordinate represents the time $t$; the vertical coordinate represents the value of our detection statistics.}
\label{fig:signals}
\vspace{-0.2in}
\end{wrapfigure}

For TEM problems, considering the property that the change usually corresponds to a sparse mean-shift (a sparse feature), we can adapt the general ACM and ASR for exponential family distributions in \cite{cao2018sequential} to the case when the signal is distributed as a Gaussian with sparse mean vector, and set the constraint set $\mathcal A$ for the unknown parameter to be $\Gamma = \{\theta: \|\theta\|_1 \leq s \}$ (refer to \cite{frecon2017bayesian} for an automatic choice of $s$) as a convex relaxation of the non-convex set that contains all sparse vectors with an upper bound on the sparsity level. Here, $\|\cdot\|_1$ and $\|\cdot\|_2$ denote the $\ell_1$ and $\ell_2$ norms in the Euclidean space, respectively.

Fig. \ref{fig:signals} gives an example of extracted signals for 15 consecutive angles; clearly, a few angles contains significant changes, which will be captured by the sequential change-point detection algorithm. $T_{\rm ACM}(b)$ and $T_{\rm ASR}(b)$ with $\Gamma = \mathbb{R}^d$ both stop at time $t=18$, CUSUM procedure with an all-one post-change mean vector stops at time $t=24$ and the generalized likelihood ratio (GLR) procedure stops at time $t=4$. Domain knowledge tells us that the change happens at time $t=17$. So the detection delays of our methods are $1$ while that of the CUSUM procedure is $7$ (meaning it takes two samples to detect). The GLR procedure raises a false alarm because it is too sensitive to noise.

\vspace{-0.1in}
\section{Spatial correlations in diffraction space}
\label{sec:spatial}
\vspace{-0.1in}
\begin{wrapfigure}{r}{0.4\textwidth}
\centering
\includegraphics[width=1.\linewidth]{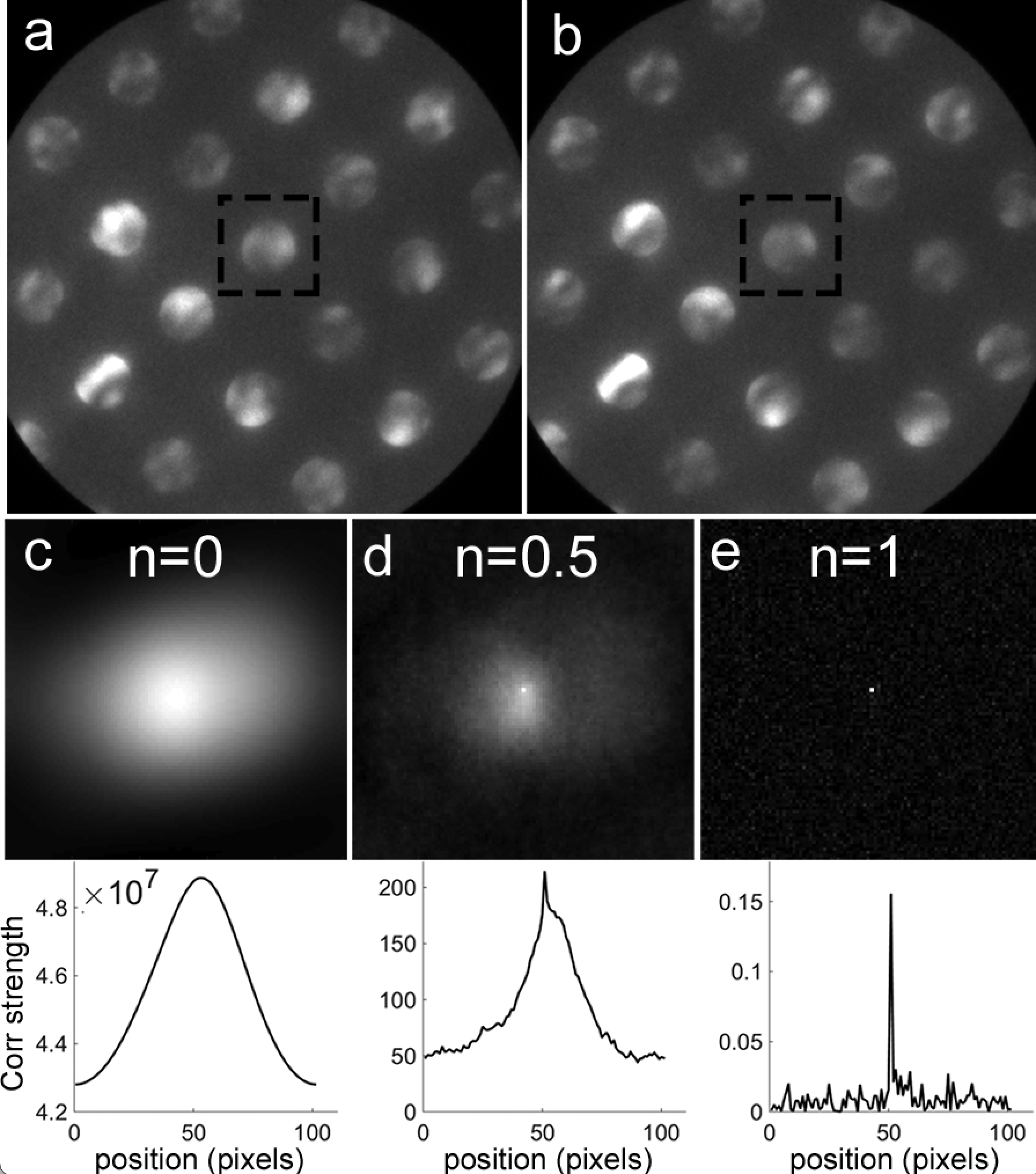}
\caption{Influence of $n$ value on peak correlations. (a, b) Nanobeam diffraction patterns taken from a PZT material. Region used for correlation indicated by the box. (c, d, e) Correlations for $n=0$, $0.5$, and $1$, respectively. Profiles below panels show the intensity distribution across the center of the field.}
\label{fig:corrphasecomp}
\vspace{-.2in}
\end{wrapfigure}

Feature tracking has long seen application in electron microscopy for purposes of drift correction. More recently, Fourier transform-based cross-correlation techniques have been applied to diffraction space to track shifts in diffraction spots or disks \cite{ozdol2015strain} \cite{savitzky2020py4dstem}. In this approach, a grid of diffraction patterns is collected from an area of interest using scanning nanobeam electron diffraction (NBED). Similar to high angular resolution electron backscatter diffraction \cite{Wilkinson2006HREBSD} \cite{Kacher2020HREBSD}, this approach relates small shifts in diffraction space to point-to-point rotations and changes in lattice parameters in real space. By collecting diffraction patterns in a grid pattern using NBED, strain gradient and crystallographic rotation maps can be produced at significantly larger scales than geometric phase analysis-based approaches and at spatial resolutions approaching 1 nm. Although the theoretical basis for relating shifts in diffraction patterns to the phase, orientation, and strain state of material has been well understood for decades, it is only with the recent advent of efficient electron detectors that nanobeam diffraction has been recognized as a viable property mapping technique. Practical applications were first demonstrated just over five years ago, with numerous materials science applications soon following, including measuring lattice parameter changes associated with chemical gradients in novel piezoelectric materials \cite{Agar2016}, tracking strains and elemental disorder associated with dislocation motion in stainless steel and Al alloys \cite{pekin2018situ,Gammer2016insitu}, and measuring strains in multilayered materials \cite{ozdol2015strain}.

Diffraction patterns are essentially a reciprocal-space representation of the probed sample volume. Thus, the spacing of diffraction disks, $g_\text{hkl}$, is equal to the inverse of the atomic plane spacing, $d_\text{hkl}$. More formally, shifts in diffraction spot spacing can be used to calculate the two-dimensional deformation gradient tensor, $\mathbf F$. This tensor can be decomposed using a polar decomposition into an elastic strain gradient component (the symmetric portion of $\mathbf F$) and a rigid body rotation tensor (the asymmetric portion of $\mathbf F$). The accuracy of the deformation gradient tensor measurements is directly related to how accurately the locations of the diffraction features (generally spots) in the diffraction patterns can be identified. As even single-pixel precision is inadequate for detecting most strains present in materials of interest, sub-pixel routines have been developed and are discussed below.



To enable rapid identification of disk locations, fast Fourier transforms (FFTs) are used to calculate the correlation between a reference disk and the NBED pattern according to:
\begin{equation}
(f * g) = \mathcal F^{-1} \left \{ \frac{\overline{\mathcal F\{f\}}\cdot \mathcal F\{g\}}{\big|\overline{\mathcal F\{f\}}\cdot \mathcal F\{g\}\big|^\gamma} \right \},
\label{eq:correlation}
\end{equation}
where $\mathcal F$ is the Fourier transform, $\star$ indicates the convolution, and $\overline X$ indicates the complex conjugate of $X$; $1-\gamma$ gives the weighting of the Fourier coefficients, with $\gamma=0$ corresponding to a standard cross-correlation and $\gamma=1$ corresponding to a phase correlation. Once calculated, the peaks in the correlation can be taken to be the location of the center of each diffraction disk. The reference disk can either be a measured disk, collected in a vacuum, or a simulated disk, based on the convergence angle and camera length used during diffraction. Under the kinematical theory of electron diffraction, the disks in the diffraction pattern all have uniform contrast, with the size of the disk determined by the convergence angle of the electron beam. However, due to the thickness and density of TEM samples, electrons usually undergo multiple scattering events, necessitating the application of dynamical theory to understand the contrast variations in the diffraction disks. This dynamic contrast varies as a function of thickness, sample bending, and localized strain field gradients (see example diffraction patterns in Fig.~\ref{fig:corrphasecomp}(a)-(b)). As all of these influences are found in most TEM samples of interest, these contrast variations must be accounted for when determining the locations of the diffraction disks. Beyond these dynamical diffraction effects, diffraction patterns also contain random noise common to all detectors. These sources of contrast fluctuations have motivated various software and hardware-oriented solutions to increasing the accuracy of the correlations.

\begin{wrapfigure}{r}{0.4\textwidth}
\centering
\includegraphics[width=1.\linewidth]{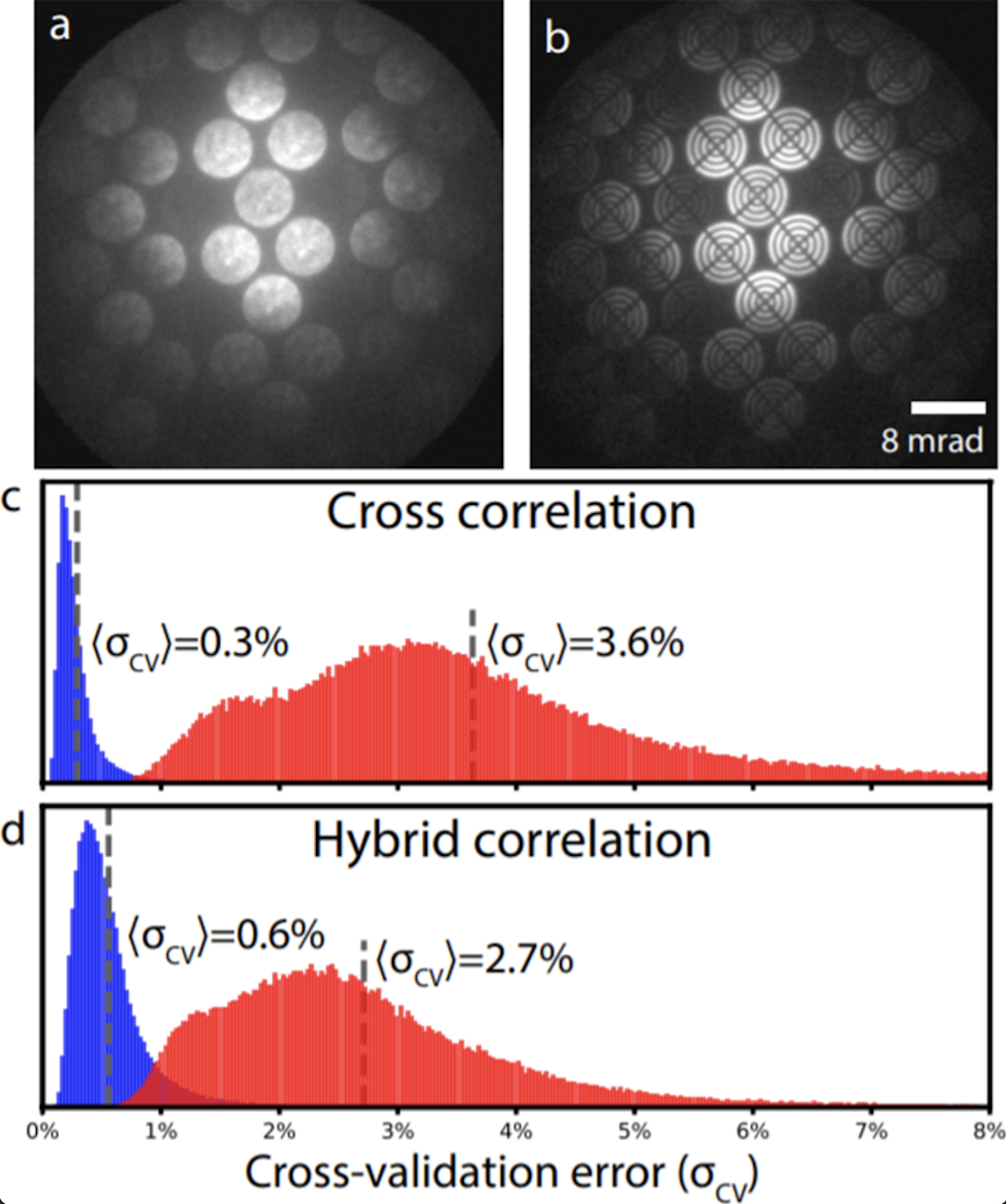}
\caption{Comparison of correlation of NBED patterns (a) with and (b) without using bullseye aperture; (c), (d): Improvements of cross correlation and hybrid correlation values when using bullseye aperture (blue distribution) compared to a standard aperature (red distribution) (figure is modified from \cite{zeltmann2019arxiv}).}
\label{fig:bullseye}
\vspace{-.2in}
\end{wrapfigure}

Fig.~\ref{fig:corrphasecomp} shows the influence of $\gamma$ in Equation \eqref{eq:correlation}. As can be seen, the peak becomes much more localized as the correlation shifts to phase correlation over standard cross-correlation. Pekin {\it et al.} explored a number of correlation and filtering-based approaches to determine the best method of determining diffraction disk location, optimizing both the accuracy and precision of the measurements as well as the computational expense \cite{pekin2017optimizing}. They found that while standard cross-correlations performed well on noisy patterns, low-frequency contrast fluctuations led to large systematic errors. In contrast, phase correlations performed well on patterns with low-frequency fluctuations but showed significant errors when high-frequency noise was present. They found that setting $\gamma=0.5$ provided the best compromise. Filtering further increased the accuracy of the correlations, especially in the presence of high-frequency noise. Using simulated and experimental patterns, they found that pre-filtering the patterns led to a significant increase in accuracy and precision of identifying diffraction spots. As NBED datasets often include tens or hundreds of thousands of distinct diffraction patterns, methods that rely on iterative processes to sharpen images and identify features such as Canny filters followed by circle fitting were ruled out. Instead, a simple Sobel filter used to sharpen edges of geometric shapes was found to improve the accuracy and precision of the correlation measurements while still being computationally efficient. Subpixel resolution in the correlations is achieved using a local upsampling approach in frequency space before fitting a parabolic curve. As the calculations are performed as matrix multiplications in the frequency domain, the refining process adds little computational expense.

While these software-based solutions improve the accuracy and precision of disk identification, which in turn improves the resolution of the elastic strain gradient measurements, the resolution is still limited to measuring relatively large elastic strains. To further increase the accuracy of disk registration, hardware solutions have been proposed and implemented, which decrease the influence of dynamical diffraction effects. One such approach is to use a precessed electron beam to average out dynamical diffraction effects, leading to quasi-kinematical diffraction and reducing the prevalence of low-frequency contrast fluctuations in the diffraction patterns \cite{cooper2016strain} \cite{vigouroux2014strain}. Rather than the electron beam entering the sample parallel to the optic axis of the microscope, the beam is rotated at a fixed angle during pattern collection. While this approach has shown improvement in disk registration, with the correlation precision approximately twice as good as achieved through standard NBED, it comes at a high cost as precession coils must be added to the microscope column. To achieve similar gains in resolution, Zeltmann {\it et al}. and Guzzinati {\it et al.} have proposed adding patterned apertures into the illumination system of the microscope \cite{zeltmann2019improved,guzzinati2019electron}. In Guzzinati  {\it et al.}’s approach, known as Bessel beam diffraction, they add a ring-shaped aperture, giving a parallel precession effect where the beam does not need to be rotated. Zeltmann {\it et al.} use a ‘bullseye’ detector to shape the beam. In both cases, the aperture pattern is passed to the diffraction disks themselves, resulting in shapes that can be readily identified using correlation algorithms. Fig.~\ref{fig:bullseye} shows a comparison between the patterns and correlation values of NBED patterns using a standard (Fig.~\ref{fig:bullseye}(a))) and bullseye ((Fig.~\ref{fig:bullseye}(b)) aperture.

\vspace{-0.1in}
\section{Future outlook}
\label{sec:conclusion}
As electron detector technologies and computational power continue to increase, the techniques described in this paper are likewise expected to improve efficiency and accuracy. However, data collection in TEM is still heavily reliant on user input. This creates a bottleneck in materials discovery due, which is increasingly being recognized as a limiting factor in materials discovery \cite{spurgeon2021towards}. TEM holders now allow precise control of local environments, including the temperature, stress state, and atmospheric conditions, facilitating {\it in-situ} material synthesis, processing, and testing at the nanoscale. However, it is often impossible to know {\it a priori} where the processes of interest will initiate, resulting in the salient phenomena occurring without observation and understanding of the transformation pathways left to speculations based on {\it post mortem} observations.

The ability to rapidly detect changes in state and accurately identify regions of interest, both in real and diffraction space, raises the possibility of overcoming the limitations associated with reliance on user input. The techniques described in this survey can be integrated directly into the microscope for in-line processing and adaptive imaging. Microscopes could be 'taught' features of interest to search for, in the case of 'needle-in-a-haystack' type problems, or detect during {\it in-situ} experiments and adjust imaging modes and resolution to optimize the data extraction process.

Further, the methods described in this paper are necessary for extracting materials knowledge from large datasets of images. The ability of researchers to efficiently analyze these images will enable the building of PSP relationships for new materials with higher accuracy due to the large datasets. This will help build the knowledge needed for understanding how material structure can be manipulated to create materials with desired properties and performance.

\section*{Acknowledgement}

The work of Shixiang Zhu, Henry Shaowu Yuchi, and Yao Xie is supported by an NSF CAREER Award CCF-1650913, CMMI-2015787,  DMS-1938106, DMS-1830210, and funding from SERDP. The work of Josh Kacher and Jordan Key is supported by the U.S. Department of Energy (DOE), Office of Science, Basic Energy Sciences (BES) Materials Science and Engineering (MSE) Division under award DE-SC0018960.

\vspace{-0.1in}
\bibliographystyle{IEEEbib}
\bibliography{refs}

\clearpage
\vspace{-0.1in}
\section*{Authors}
All authors are affiliated with the Georgia Institute of Technology. 
\begin{itemize}
\item Yao Xie is an associate professor in the School of Industrial and Systems Engineering. Her research includes statistical signal processing and machine learning. She receives the NSF CAREER Award in 2017. She is currently an Associate Editor for IEEE Transactions on Signal Processing, Sequential Analysis: Design Methods and Applications, INFORMS Journal on Data Science, and serves on the Editorial Board of Journal of Machine Learning Research.
\item Josh Kacher is an assistant professor in the School of Materials Science and Engineering at Georgia Institute of Technology. He is the recipient of the 2020 NSF CAREER Award, the 2015 DOE Early Career award, and the 2017 ONR Young Investigator award. His research focuses on understanding material behavior under a wide range of environmental conditions.
\item Surya Kalidindi is a Regents Professor, and Rae S. and Frank H. Neely Chair Professor in the Woodruff School of Mechanical Engineering at Georgia Institute of Technology, Georgia, USA with joint appointments in the School of Materials Science and Engineering as well as the School of Computational Science and Engineering. Dr. Kalidindi's research interests are broadly centered on designing material internal structure (including composition) for optimal performance in any selected application and identifying hybrid processing routes for its manufacture. To this end, he has employed a harmonious blend of experimental, theoretical, and numerical approaches in his research.
\item Shixiang Zhu
received the B.S. and M.S. degree in computer science from Beijing University of Posts and Telecommunications in 2014 and 2017, respectively. He is currently pursuing the Ph.D. degree in Machine Learning in H. Milton Stewart School of Industrial and Systems Engineering, Georgia Institute of Technology. His research focus on machine learning and data science inspired by important applications, including police operation, power grid resilience, intelligent transportation, engineering seismology, financial security, healthcare, and diffraction and imaging in material science.
\item Sven Voigt received his M.S. in Materials Science from CWRU in 2018, and interned at Novelis Inc as a data engineer. He joined the Materials Science Graduate program at Georgia Institute of Technology as an NSF FLAMEL trainee and is working on developing a materials knowledge infrastructure. 
\item Henry Shaowu Yuchi is a PhD student in Machine Learning student in the School of Industrial and Systems Engineering; he received the BA in Engineering and MEng at University of Cambridge, UK. 
\item Jordan W. Key received B.S. degrees in Mechanical Engineering and Physics from the University of Arkansas in 2015 before joining the Materials Science and Engineering program at Georgia Tech. There he specialized in combining rapid data-driven analysis with novel experimental methods to study corrosion resistance, receiving his Ph.D. in 2020.
\end{itemize}

\end{document}